\documentclass[11pt]{article}

\usepackage[final]{acl}

\usepackage{times}
\usepackage{latexsym}

\usepackage[T1]{fontenc}

\usepackage{amsmath}
\usepackage{amssymb}
\usepackage[utf8]{inputenc}

\usepackage{microtype}
\usepackage{inconsolata}
\usepackage{graphicx}
\usepackage{booktabs}
\usepackage{multirow}
\usepackage{xcolor}
\usepackage{subcaption}
\usepackage{tikz}
\usetikzlibrary{shapes.geometric, arrows.meta, positioning, calc, fit}

\title{Probing Ambiguity: Adversarial Intent Resolution for Robust Text-to-SQL}
\author{
   Youbing Yin, Xujun Peng, Muhammad S. Uddin, Erin Babinsky, Daben Liu, Anoop Kumar  \\
  Capital One \\ \texttt{\{youbing.yin, xujun.peng, muhammad.uddin,} \\ \texttt{erin.babinsky, daben.liu, anoop.kumar\}@capitalone.com}
}

\begin{document}
\maketitle
\begin{abstract}
Text-to-SQL systems typically assume a one-to-one mapping between natural-language queries and correct SQL translations. In practice, however, many queries are inherently ambiguous,  admitting multiple valid interpretations that yield distinct outputs. We introduce Adversarial Intent Resolution (AIR), a multi-agent framework that actively probes and resolves such ambiguity without any task-specific training. AIR first constructs competing interpretations, verifies through SQL execution that they produce genuinely different results, and generates targeted clarification questions grounded in the observed data differences. Evaluating on leading benchmarks for ambiguous questions, AIR demonstrates strong performance, efficiency, and robust cross-benchmark transferability. On the AMBROSIA benchmark under schema-only context—a realistic production setting where exposing the full database is infeasible—AIR outperforms IntentRL, the state-of-the-art reinforcement learning model, across all metrics (e.g., F1 0.679 vs. 0.618). Crucially, AIR exhibits only $\sim$7\% performance degradation without the full database, compared to $\sim$20\% for IntentRL.  On the AmbiQT benchmark, where ambiguity arises from schema design rather than linguistic structure, AIR achieves higher F1 under both context settings without any benchmark-specific adaptation.

\end{abstract}

\section{Introduction}

Text-to-SQL systems have made remarkable progress, with recent LLM-based approaches achieving over 85\% execution accuracy on benchmarks like Spider \citep{yu2018spider, lei2025spider} and BIRD \cite{li2024bird} through in-context learning \cite{guo2024dail, pourreza2024din} and fine-tuning \cite{li2024codes}. These advances bring closer the promise of natural-language database interfaces, enabling non-technical users to query complex schema without writing SQL. Yet this progress rests on a critical assumption: \emph{that each natural-language question maps to exactly one correct SQL translation}.

In practice, this assumption is fundamentally at odds with how humans formulate questions. Consider: ``Show \$150 wellness programs and day spas.'' Should the price filter apply to both wellness programs and day spas, or only wellness programs? Under one interpretation the answer is two rows; under the other, six rows, a very different result set.  This is not an edge case. Ambiguous queries routinely admit multiple valid SQL translations that return different results \cite{bhaskar-etal-2023-benchmarking, saparina2025ambrosia}.

Yet the dominant paradigm maps each question to a single SQL query and declares success \cite{yao2025arctic, singh2026comprehensive, liu2026systematic}. Even when systems are explicitly prompted for diverse outputs, they overwhelmingly generate surface-level variants of the top prediction rather than structurally distinct alternatives \cite{bhaskar-etal-2023-benchmarking}. Diversity-promoting decoding strategies such as beam search and nucleus sampling fail to explore genuinely different interpretations. A recent work trains models via reinforcement learning (RL) to generate multiple candidate interpretations \cite{saparina2026reasoningintentambiguousrequests}, but still treats ambiguity as a generation problem rather than a \emph{probing and verification} problem. Moreover, in realistic production settings the full database may not be available as the context to LLM due to privacy, latency, and cost constraints, restricting the LLM to schema-level context only. It remains unclear how these approaches perform when schema-only is used as context to guide disambiguation.

We propose \textbf{Adversarial Intent Resolution (AIR)}, a multi-agent framework that requires no task-specific training and resolve ambiguity through \emph{adversarial probing}, actively attempting to construct a competing interpretation. An Anchor agent proposes the most likely interpretation; an Adversary agent, constrained to diverge, actively constructs alternatives; and execution-based verification retains only interpretation pairs that produce different data---constituting a constructive proof of ambiguity.

We evaluate AIR on AMBROSIA \cite{saparina2025ambrosia} and AmbiQT \cite{bhaskar-etal-2023-benchmarking}. On AMBROSIA, under schema-only context, the realistic production setting, AIR outperforms IntentRL \cite{saparina2026reasoningintentambiguousrequests}, the RL-trained state-of-the-art, across all metrics (e.g., F1 0.679 vs.\ 0.618). AIR degrades only one-third as much as IntentRL when moving from full database to schema-only context.  With full database available, AIR achieves competitive recall (0.818 vs.\ 0.854) while generating fewer outputs (1.76 vs.\ 5.00 per query). On AmbiQT, AIR achieves higher F1 without any benchmark-specific adaptation.

Our main contributions are:
\begin{enumerate}
    \item We develop a multi-agent framework that uncovers ambiguity through adversarial probing, verifies it through execution-based divergence, and resolves it through contrastive clarification, requiring no task-specific training.
    \item We show that AIR outperforms the RL-trained state-of-the-art on all metrics under schema-only context, the realistic production setting, and more crucially, with only one-third the degradation.
    \item We demonstrate transferability to new ambiguity types without explicit benchmark-specific adaptation.
\end{enumerate}

\definecolor{phaseblue}{RGB}{50,100,180}
\definecolor{phaseorange}{RGB}{200,120,50}
\definecolor{phasegreen}{RGB}{60,130,80}
\definecolor{phasebluelight}{RGB}{237,243,255}
\definecolor{phaseorangelight}{RGB}{255,247,237}
\definecolor{phasegreenlight}{RGB}{237,252,241}
\definecolor{agentborder}{RGB}{70,110,200}
\definecolor{agentfill}{RGB}{220,232,255}
\definecolor{resultborder}{RGB}{120,120,120}
\definecolor{resultfill}{RGB}{248,248,248}
\definecolor{orangediamond}{RGB}{200,130,40}
\definecolor{orangediamondfill}{RGB}{255,240,210}
\definecolor{sqlblue}{RGB}{215,232,255}
\definecolor{execorange}{RGB}{255,232,215}
\definecolor{clarifgreen}{RGB}{215,248,225}
\definecolor{redarrow}{RGB}{190,60,60}
\definecolor{redlabel}{RGB}{180,50,50}
\definecolor{queryfill}{RGB}{255,253,235}
\definecolor{grayfill}{RGB}{238,238,238}
\definecolor{darkgray}{RGB}{70,70,70}

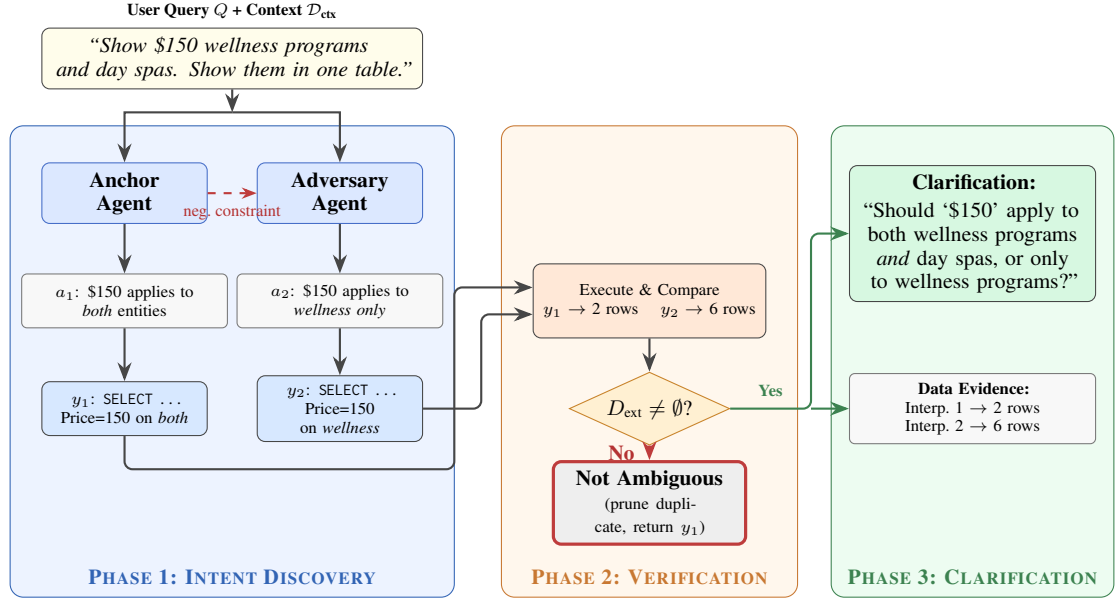
\begin{figure*}[t]
\centering
\begin{tikzpicture}[
    scale=0.89, transform shape,
    every node/.style={font=\small},
    agent/.style={rectangle, draw=agentborder, fill=agentfill, rounded corners=3pt,
                  minimum height=0.9cm, align=center, font=\small\bfseries, text width=2.2cm},
    interp/.style={rectangle, draw=resultborder, fill=resultfill, rounded corners=2pt,
                   minimum height=0.8cm, align=center, font=\scriptsize, text width=2.8cm},
    decision/.style={diamond, draw=orangediamond, fill=orangediamondfill, aspect=2.2,
                     minimum width=2.2cm, align=center, font=\small, inner sep=2pt},
    arrow/.style={-{Stealth[length=2.5mm]}, thick, draw=darkgray},
    sqlbox/.style={rectangle, draw=darkgray, fill=sqlblue, rounded corners=3pt,
                   minimum height=0.8cm, align=center, font=\scriptsize, text width=2.2cm},
    execbox/.style={rectangle, draw=darkgray, fill=execorange, rounded corners=3pt,
                   minimum height=0.8cm, align=center, font=\scriptsize, text width=3.2cm},
]

% ============================================================
% Layout: Rows aligned across phases for simple horizontal arrows.
%   Row A (y=2.8): Agents
%   Row B (y=1.2): Interpretations | Exec box        | Clarification
%   Row C (y=-0.4): SQL boxes      | Decision        | Evidence
%   Row D (y=-1.8): (empty)        | Prune           | (empty)
% ============================================================

% === PHASE BACKGROUNDS ===
\node[rectangle, rounded corners=5pt, draw=phaseblue, fill=phasebluelight,
      minimum width=6.6cm, minimum height=7.0cm] (p1bg) at (0, 0.3) {};
\node[rectangle, rounded corners=5pt, draw=phaseorange, fill=phaseorangelight,
      minimum width=4.4cm, minimum height=7.0cm] (p2bg) at (6.2, 0.3) {};
\node[rectangle, rounded corners=5pt, draw=phasegreen, fill=phasegreenlight,
      minimum width=4.2cm, minimum height=7.0cm] (p3bg) at (11.0, 0.3) {};

% === PHASE LABELS (bottom) ===
\node[font=\footnotesize\bfseries, color=phaseblue, anchor=south] at ([yshift=2pt]p1bg.south) {\textsc{Phase 1: Intent Discovery}};
\node[font=\footnotesize\bfseries, color=phaseorange, anchor=south] at ([yshift=2pt]p2bg.south) {\textsc{Phase 2: Verification}};
\node[font=\footnotesize\bfseries, color=phasegreen, anchor=south] at ([yshift=2pt]p3bg.south) {\textsc{Phase 3: Clarification}};

% === INPUT QUERY ===
\node[rectangle, draw=darkgray, fill=queryfill, rounded corners=3pt,
      minimum height=0.9cm, text width=5.4cm, align=center, font=\small] (query) at (0, 4.8) {%
      \textit{``Show \$150 wellness programs}\\\textit{and day spas. Show them in one table.''}};
\node[font=\scriptsize\bfseries, anchor=south] at (query.north) {User Query $Q$ + Context $\mathcal{D}_{\text{ctx}}$};

% ============ PHASE 1 ============

% Row A: Agents
\node[agent] (anchor) at (-1.6, 2.8) {Anchor\\Agent};
\node[agent] (adversary) at (1.6, 2.8) {Adversary\\Agent};

% Query → Agents
\draw[arrow] (query.south) -- (0, 4.0);
\draw[arrow] (0, 4.0) -- (-1.6, 4.0) -- (anchor.north);
\draw[arrow] (0, 4.0) -- (1.6, 4.0) -- (adversary.north);

% Neg constraint
\draw[-{Stealth[length=2mm]}, dashed, thick, draw=redarrow]
    (anchor.east) -- node[below, font=\scriptsize, color=redlabel, yshift=-1pt] {neg.\ constraint} (adversary.west);

% Row B: Interpretations
\node[interp] (a1) at (-1.6, 1.2) {$a_1$: \$150 applies to\\ \emph{both} entities};
\node[interp] (a2) at (1.6, 1.2) {$a_2$: \$150 applies to\\ \emph{wellness only}};

\draw[arrow] (anchor.south) -- (a1.north);
\draw[arrow] (adversary.south) -- (a2.north);

% Row C: SQL boxes
\node[sqlbox] (sql1) at (-1.6, -0.4) {$y_1$: \texttt{SELECT ...}\\Price=150 on \emph{both}};
\node[sqlbox] (sql2) at (1.6, -0.4) {$y_2$: \texttt{SELECT ...}\\Price=150 on \emph{wellness}};

\draw[arrow] (a1.south) -- (sql1.north);
\draw[arrow] (a2.south) -- (sql2.north);

% ============ PHASE 2 ============

% Row B: Execute & Compare (same row as interpretations → horizontal entry)
\node[execbox, minimum height=1.1cm] (exec) at (6.2, 1.2) {%
      Execute \& Compare\\
      $y_1 \rightarrow$ 2 rows \quad $y_2 \rightarrow$ 6 rows};

% Row C: Decision diamond (same row as SQL boxes)
\node[decision] (div) at (6.2, -0.4) {$D_{\text{ext}} \neq \emptyset$?};

\draw[arrow] (exec.south) -- (div.north);

% Row D: Prune — made visually prominent to emphasize the non-ambiguous path
\node[rectangle, draw=redarrow, fill=grayfill, rounded corners=3pt, line width=1.2pt,
      minimum height=0.9cm, text width=2.6cm, align=center, font=\small\bfseries] (prune) at (6.2, -1.8) {Not Ambiguous\\{\normalfont\scriptsize(prune duplicate, return $y_1$)}};
\draw[arrow, draw=redarrow, line width=1.4pt] (div.south) -- node[left, font=\small\bfseries, color=redlabel, xshift=-2pt] {No} (prune.north);

% ============ PHASE 3 ============

% Row B: Clarification (same row as exec)
\node[rectangle, draw=darkgray, fill=clarifgreen, rounded corners=3pt,
      minimum height=1.6cm, text width=3.4cm, align=center, font=\small] (clarify) at (11.0, 2.2) {%
      \textbf{Clarification:}\\[2pt]
      ``Should `\$150' apply to\\
      both wellness programs\\
      \emph{and} day spas, or only\\
      to wellness programs?''};

% Row C/D: Evidence
\node[interp, text width=3.4cm] (evidence) at (11.0, -0.4) {%
      \textbf{Data Evidence:}\\
      Interp.\ 1 $\rightarrow$ 2 rows\\
      Interp.\ 2 $\rightarrow$ 6 rows};

% ============ CROSS-PHASE ARROWS ============

% Phase 1 → Phase 2: SQL boxes → Exec box
% Both arrows route through the gap between Phase 1 and Phase 2.
% They use SEPARATE vertical channels (x=3.3 and x=3.7) so they don't merge.

% sql1: exit bottom, go below sql row, right to channel x=3.3, up to exec upper port
\draw[arrow, rounded corners=3pt] (sql1.south) -- (-1.6, -1.3) -- (3.3, -1.3) -- (3.3, 1.4) -- (exec.west |- 0,1.4);

% sql2: exit right, go to channel x=3.7, up to exec lower port
\draw[arrow, rounded corners=3pt] (sql2.east) -- (3.7, -0.4) -- (3.7, 1.0) -- (exec.west |- 0,1.0);

% Phase 2 → Phase 3: Decision "Yes" → Clarification and Evidence
\draw[arrow, draw=phasegreen, rounded corners=3pt] (div.east) --
    node[above, font=\scriptsize\bfseries, color=phasegreen, yshift=1pt] {Yes}
    (8.6, -0.4) -- (8.6, 2.2) -- (clarify.west);
\draw[arrow, draw=phasegreen, rounded corners=3pt] (8.6, -0.4) -- (evidence.west);

\end{tikzpicture}
\caption{Overview of the Adversarial Intent Resolution (AIR) framework. Given an ambiguous query, the Anchor agent proposes a primary interpretation while the Adversary agent---constrained to diverge---discovers an alternative. Both are compiled to SQL and executed against the database. If the result sets differ ($D_{\text{ext}} \neq \emptyset$), a contrastive clarification grounded in actual data differences is presented to the user for further clarification; otherwise the duplicate interpretation is pruned.}
\label{fig:framework}
\end{figure*}

\section{Related Work}

\paragraph{Ambiguity in Text-to-SQL.}
Recent benchmarks have formalized the ambiguity problem: AmbiQT \cite{bhaskar-etal-2023-benchmarking} pairs ambiguous questions with multiple gold SQL queries arising from schema-level ambiguities, and AMBROSIA \cite{saparina2025ambrosia} covers scope, attachment, and vagueness ambiguities with execution-based evaluation. To address ambiguity, \citet{saparina2025disambiguate} separates disambiguation from SQL generation by training an infilling model to recover missed interpretations, and \citet{saparina2026reasoningintentambiguousrequests} uses RL to generate multiple interpretations in a single forward pass. A common limitation is their \emph{passive} treatment: candidates are generated but no method actively constructs a competing interpretation and verifies, through execution, that it produces different data. AIR shares the principle of separating disambiguation from SQL generation \cite{saparina2025disambiguate}, but replaces trained components with adversarial probing---using negative constraints to force exploration of the interpretation space.

\paragraph{Clarification Question Generation.}
Several systems address ambiguity by generating clarifying questions. \citet{ding2026ambisqlinteractiveambiguitydetection} designed a framework to identify and resolve ambiguities in Text-to-SQL tasks by engaging users in a multi-turn dialogue to clarify their intent before generating the final database query. \citet{zhang2024reactable} explored interactive clarification with LLMs in broader semantic parsing tasks. These approaches generate clarifications based on perceived ambiguity without first proving that different interpretations lead to different data outcomes. AIR generates clarifications only after execution-based verification confirms genuine divergence, reducing unnecessary user interruptions.

\paragraph{Multi-Agent and Adversarial LLM Frameworks.}
Multi-agent architectures have shown promise for improving reasoning quality. \citet{du2024debate} demonstrated that multi-agent debate improves factuality, and in Text-to-SQL, \citet{wang2024macsql} proposed MAC-SQL, where specialized agents handle decomposition, generation, and refinement. These frameworks share a \emph{convergent} objective: agents iteratively improve a single answer.  \citet{liang2024encouraging} instead uses structured disagreement to encourage divergent thinking.  However, in AIR, the Adversary agent seeks to falsify the Anchor's uniqueness, forcing exploration of the interpretation space rather than convergence toward a single answer.  Adversary must diverge from one concrete anchor interpretation rather than disagreeing in general, and execution-based verification proves the two interpretations are genuinely distinct rather than relying on disagreement alone.

\section{The Adversarial Intent Resolution Framework}
The Adversarial Intent Resolution (AIR) framework treats Text-to-SQL ambiguity as a \emph{probing and verification} problem. Without any task-specific training, AIR actively constructs competing interpretations through adversarial probing, verifies through SQL execution that they produce different results, and generates targeted clarification questions grounded in the observed data differences. 

\subsection{Problem Formulation}
Given a natural-language query $Q$, a database $\mathcal{D}$, and a context $\mathcal{D}_{\text{ctx}}$ provided to the LLM (which may be the full database or only the schema), a standard Text-to-SQL system produces a single SQL query $\hat{y}$ maximizing $P(y \mid Q, \mathcal{D}_{\text{ctx}})$. We characterize the set of valid interpretations $\mathcal{I}(Q, \mathcal{D}_{\text{ctx}})$ as a collection of pairs $(a_i, y_i)$, where each $a_i$ is a natural-language interpretation and $y_i$ is its corresponding SQL translation. 

Intent discovery and SQL generation operate on $\mathcal{D}_{\text{ctx}}$ (which may be schema-only in production), while execution-based verification runs against the full database $\mathcal{D}$.

AIR targets {extensional} ambiguity: a query is ambiguous when at least two valid interpretations produce different observable results on $\mathcal{D}$, i.e., $\exists\, i \neq j$ with $\text{Exec}(y_i, \mathcal{D}) \neq \text{Exec}(y_j, \mathcal{D})$. Two interpretations that differ in wording but happen to return the same result on the current data are not flagged, since the ambiguity would not change the answer.

\subsection{Phase 1: Adversarial Intent Discovery}

AIR employs a multi-agent architecture where agents reason exclusively at the intent level, producing natural-language interpretations before generating executable SQL.

\paragraph{Anchor Agent.}
The Anchor agent produces the primary interpretation:
\begin{equation}
    (a_1, c_1) = \textsc{Anchor}(Q, \mathcal{D}_{\text{ctx}})
\end{equation}
where $a_1$ is a natural-language interpretation stating which ambiguous elements were resolved and how, and $c_1 \in [0, 1]$ is a confidence score indicating this is the only reasonable interpretation.

\paragraph{Adversary Agent.}
The Adversary receives the Anchor's interpretation $a_1$ as a negative constraint and attempts to find an alternative:
\begin{equation}
    (a_2, c_2, f) = \textsc{Adversary}(Q, \mathcal{D}_{\text{ctx}}, a_1)
\end{equation}
where $f \in \{\texttt{true}, \texttt{false}\}$ indicates whether a valid alternative was found, with a confidence score of $c_2 \in [0, 1]$.

\paragraph{Confidence-Based Gating.}
AIR applies confidence-based gating at two stages: (1) if $c_1 \geq \tau_{\text{anchor}}$ (default $\tau_{\text{anchor}} = 0.95$), the Adversary is suppressed entirely; (2) if the Adversary finds an alternative but with $c_2 < \tau_{\text{adv}}$  (default $\tau_{\text{adv}} = 0.5$), the alternative is discarded as insufficiently plausible.

\paragraph{Multi-Round Adversary.}
For queries with more than two valid interpretations, the Adversary is called iteratively: after discovering $a_2$, it receives all previously discovered interpretations $\{a_1, a_2, \ldots\}$ as negative constraints and searches for yet another distinct reading. This continues until no new alternative is found or a maximum of $K$ interpretations is reached (default $K = 5$). 

\subsection{Phase 2: Extensional Divergence Verification}

Different interpretations may yield identical query results, so language-level differences alone do not constitute genuine ambiguity. AIR verifies divergence by compiling each interpretation to SQL and executing it.

\paragraph{SQL Generation.}
For each interpretation $a_i$, a SQL generation agent translates it into executable SQL:
\begin{equation}
    y_i = \textsc{SQLGen}(Q, \mathcal{D}_{\text{ctx}}, a_i)
\end{equation}

This step operates with an execution feedback loop: if the generated SQL produces a database error, it receives the error message and retries (up to 3 attempts).  

\paragraph{Execution-Based Pruning.}
For each pair of interpretations, we compute the symmetric difference of their result sets:
\begin{equation}
    D_{\text{ext}}(y_i, y_j) = \text{Exec}(y_i, \mathcal{D}) \;\Delta\; \text{Exec}(y_j, \mathcal{D})
\end{equation}
If $D_{\text{ext}} = \emptyset$, the two interpretations are equivalent on the current data and one is pruned. Only interpretations that produce results distinct from all others are retained as genuinely ambiguous.

\subsection{Phase 3: Contrastive Clarification Synthesis}

When genuine ambiguity is confirmed, a Synthesizer agent generates a clarification grounded in the actual data differences:
\begin{equation}
    r = \textsc{Synth}(Q, \{(a_i, \text{Exec}(y_i, \mathcal{D}))\}_{i=1}^{n})
\end{equation}
where $n$ is the number of retained interpretations after pruning. The Synthesizer receives each interpretation paired with its execution results, enabling it to produce \emph{contrastive} clarifications that highlight how results differ under each interpretation, rather than asking generic clarification questions.

\subsection{Few-Shot Prompting}
\label{subsec:few-shot}
The intent discovery and SQL generation agents share a common set of five few-shot examples drawn from AMBROSIA's few-shot example split \citep{saparina2025ambrosia}, covering the three ambiguity types (scope, attachment, vagueness) plus a non-ambiguous control. These examples help teach the agents the structural patterns that create each ambiguity type and their corresponding SQL translations.

\subsection{Illustrative Example}

Figure~\ref{fig:example} shows AIR applied to an attachment ambiguity query from AMBROSIA.

\begin{figure*}[t]
\centering
\small
\fbox{\parbox{\textwidth}{
\textbf{Query:} \textit{``Show \$150 wellness programs and day spas. Show them in one table.''}

\medskip
\textbf{Phase 1: Adversarial Intent Discovery}

\textit{Anchor Agent:} ``\$150'' modifies the conjunction ``wellness programs and day spas'' $\to$ filter \emph{both} tables by Price = 150.

\textit{Adversary Agent} (receives Anchor as negative constraint): ``\$150'' attaches only to ``wellness programs'' $\to$ filter wellness programs by price, but include \emph{all} day spas.

\medskip
\textbf{Phase 2: Extensional Divergence Verification}

\begin{tabular}{@{}p{0.45\textwidth}p{0.45\textwidth}@{}}
\textit{SQL 1 (Anchor):} & \textit{SQL 2 (Adversary):} \\
\texttt{\small SELECT Name FROM Wellness\_Retreat} & \texttt{\small SELECT Name FROM Wellness\_Retreat} \\
\texttt{\small ~WHERE Price = 150} & \texttt{\small ~WHERE Price = 150} \\
\texttt{\small UNION} & \texttt{\small UNION} \\
\texttt{\small SELECT Name FROM Day\_Spa} & \texttt{\small SELECT Name FROM Day\_Spa} \\
\texttt{\small ~WHERE Price = 150} & \\
\end{tabular}

\smallskip
Execution results differ: SQL 1 returns \{Mountain Retreat, Oasis Day Spa\} (2 rows); SQL 2 returns \{Mountain Retreat, Oasis Day Spa, Escape Day Spa, Paradise Day Spa, Serenity Day Spa, Zen Day Spa\} (6 rows). $D_{\text{ext}} \neq \emptyset$ $\to$ ambiguity confirmed, both retained.

\medskip
\textbf{Phase 3: Contrastive Clarification Synthesis}

\textit{``I found two possible interpretations:}\\
\textit{1. Both wellness programs AND day spas that cost \$150} $\to$ \textit{Mountain Retreat, Oasis Day Spa (2 results)}\\
\textit{2. Wellness programs at \$150, plus all day spas regardless of price} $\to$ \textit{6 results}\\
\textit{Which did you mean?''}
}}
\caption{AIR correctly identifies an attachment ambiguity: does ``\$150'' modify both conjuncts or only ``wellness programs''? The adversarial probe finds the alternative interpretation, execution confirms divergent outputs (2 vs.\ 6 rows), and the synthesizer presents a contrastive clarification grounded in actual data differences.}

\label{fig:example}
\end{figure*}

\section{Experimental Setup}
\subsection{Datasets}
\textbf{AMBROSIA}. AMBROSIA~\cite{saparina2025ambrosia} is a Text-to-SQL benchmark where natural-language questions exhibit linguistic ambiguity (scope, attachment, vagueness), leading to multiple valid SQL translations with different execution results. Each query is paired with gold interpretations and a SQLite database enabling execution-based evaluation. We use the data split from \citet{saparina2026reasoningintentambiguousrequests}: 131 ambiguous and 309 unambiguous test examples (440 total).

\textbf{AmbiQT}. To assess cross-benchmark generalization, we evaluate on AmbiQT \cite{bhaskar-etal-2023-benchmarking}. On this benchmark, ambiguity originates from schema design rather than linguistic structure: real-world Spider \cite{yu2018spider} databases are augmented with schema-level redundancies (synonym columns, synonym tables, normalized splits, pre-computed aggregates) so that a single question admits multiple structurally valid SQL translations. Following \citet{saparina2026reasoningintentambiguousrequests}, we use the preprocessed validation split.

\subsection{Context Conditions}

We evaluate under two context settings for $\mathcal{D}_{\text{ctx}}$:
\begin{itemize}
    \item \textbf{Full database:} $\mathcal{D}_{\text{ctx}} = \mathcal{D}$, the system receives the complete database (schema plus all data) as the context
    \item \textbf{Schema only:} $\mathcal{D}_{\text{ctx}}$ contains only the schema of the database, with no data.
\end{itemize}
The schema-only setting simulates production scenarios where exposing actual data to the LLM is infeasible due to privacy, scale, or latency constraints. We emphasize that ``schema-only'' describes what is passed to the LLM as context, not whether the system can query the database: SQL execution (Phase~2) and the Synthesizer (Phase~3) still run against the full database $\mathcal{D}$ as a backend operation under standard database access controls, identical to any ordinary application query. The privacy, cost, and latency concern is specific to sending database content to a third-party LLM as context.

\subsection{Implementation Details}
We evaluate AIR with three models representing diverse architectures: Claude Sonnet 4.5 \citep{anthropic2025claude}, Gemma 4 26B A4B \citep{team2026gemma}, and gpt-oss-20b \citep{openai2025gptoss}. All configurations use temperature 0 and their default reasoning capabilities. No task-specific training or fine-tuning is performed for any configuration.

\subsection{Baselines}

\paragraph{Prompt Baseline.}
We establish a prompting baseline by directly instructing the LLM to generate multiple SQL interpretations for each query. We evaluate this with Claude Sonnet 4.5.

\paragraph{IntentRL.}

Our primary comparison is IntentRL \cite{saparina2026reasoningintentambiguousrequests}, the current state-of-the-art on AMBROSIA. IntentRL is an RL model fine-tuned on AMBROSIA's training split, using a composite reward for ambiguity detection and multi-interpretation SQL generation. It generates 5 SQL queries per question in a single forward pass. We select IntentRL over DisambigParse \cite{saparina2025disambiguate}, a two-stage method that first predicts interpretations then generates SQL. On ambiguous queries, IntentRL outperforms DisambigParse across all metrics (e.g., recall 82.4\% vs.\ 59.7\%, precision 77.5\% vs.\ 26.3\%). It also surpasses prompting with a 235B model (46.2\% recall) and supervised fine-tuning (68.5\% recall). This makes it the most stringent test of whether inference-time adversarial reasoning can match training-based approaches.

\subsection{Evaluation Metrics}
Following \citet{saparina2025ambrosia}, we use execution-based evaluation metrics. Since ambiguous queries map to multiple correct SQL translations, we report recall, precision, and F1.  We also report One-Found (at least one gold execution output covered by predictions) and All-Found (all gold execution outputs covered by predictions).

\section{Results}
\subsection{AMBROSIA Results}

\begin{table*}[t]
\centering
\small
\setlength{\tabcolsep}{4.5pt}
\begin{tabular}{ll ccccc}
\toprule
\textbf{Context} & \textbf{Method} & \textbf{Recall} & \textbf{Precision} & \textbf{F1} & \textbf{One-Found} & \textbf{All-Found} \\
\midrule
\multirow{5}{*}{Full database}
 & Prompt Baseline (Sonnet 4.5) & 0.697 & 0.592 & 0.610 & 0.797 & 0.590 \\
 & SOTA (IntentRL)              & \textbf{0.854} & \textbf{0.758} & \textbf{0.786} & \textbf{0.886} & \textbf{0.823} \\
\addlinespace[2pt]
 & AIR -- Claude Sonnet 4.5 (ours)    & 0.818 & 0.707 & 0.729 & 0.880 & 0.758 \\
 & AIR -- Gemma 4 26B A4B (ours)   & 0.804 & 0.584 & 0.640 & 0.876 & 0.729 \\
 & AIR -- gpt-oss-20b (ours)       & 0.759 & 0.728 & 0.718 & 0.845 & 0.673 \\
\addlinespace[6pt]
\multirow{5}{*}{Schema only}
 & Prompt Baseline (Sonnet 4.5) & 0.610 & 0.539 & 0.548 & 0.698 & 0.520 \\
 & SOTA (IntentRL)              & 0.691\,{\scriptsize($-$19\%)} & 0.589\,{\scriptsize($-$22\%)} & 0.618\,{\scriptsize($-$21\%)} & 0.723\,{\scriptsize($-$18\%)} & 0.664\,{\scriptsize($-$19\%)} \\
\addlinespace[2pt]
 & AIR -- Claude Sonnet 4.5 (ours)    & \textbf{0.757}\,{\scriptsize($-$7\%)} & \textbf{0.658}\,{\scriptsize($-$7\%)} & \textbf{0.679}\,{\scriptsize($-$7\%)} & \textbf{0.813}\,{\scriptsize($-$8\%)} & \textbf{0.701}\,{\scriptsize($-$8\%)} \\
 & AIR -- Gemma 4 26B A4B (ours)   & 0.750\,{\scriptsize($-$7\%)} & 0.518\,{\scriptsize($-$11\%)} & 0.578\,{\scriptsize($-$10\%)} & 0.820\,{\scriptsize($-$6\%)} & 0.682\,{\scriptsize($-$6\%)} \\
 & AIR -- gpt-oss-20b (ours)       & 0.648\,{\scriptsize($-$15\%)} & 0.642\,{\scriptsize($-$12\%)} & 0.626\,{\scriptsize($-$13\%)} & 0.728\,{\scriptsize($-$14\%)} & 0.568\,{\scriptsize($-$16\%)} \\
\bottomrule
\end{tabular}
\caption{Performance on AMBROSIA test split. Percentages indicate degradation from full database to schema-only context. IntentRL leads under full database context, but under schema-only, the realistic production setting, AIR (Claude Sonnet 4.5) outperforms IntentRL across all metrics with roughly one-third the degradation (e.g., F1: $\sim$7\% vs.\ $\sim$20\%). AIR with smaller open-source models also exhibits less degradation than IntentRL, confirming that the robustness stems from the framework architecture.}
\label{tab:overall}
\end{table*}

\subsubsection{Performance with Schema-Only as Context}
In production deployments, exposing full database contents to an LLM is often infeasible due to privacy, latency, and cost constraints. A training-free approach that operates effectively with schema-only access is therefore highly desirable. Table~\ref{tab:overall} presents performance under this realistic setting. AIR (Claude Sonnet 4.5) outperforms IntentRL across all metrics: recall (0.757 vs.\ 0.691), precision (0.658 vs.\ 0.589), F1 (0.679 vs.\ 0.618), One-Found (0.813 vs.\ 0.723), and All-Found (0.701 vs.\ 0.664). Taking F1 as an example, this represents a 10\% relative improvement, demonstrating that AIR's adversarial probing architecture produces effective disambiguation even without access to the actual data.

Even with smaller open-source models, AIR outperforms IntentRL for the schema-only setting (F1: gpt-oss-20b 0.626 vs.\ IntentRL's 0.618), validating that AIR's advantage stems from the framework rather than just the backbone model.

Two further comparisons in Table~\ref{tab:overall} isolate the framework's contribution from backbone model capacity. First, the Prompt Baseline uses the same Claude Sonnet 4.5 backbone as AIR but without the adversarial structure, scoring schema-only F1 0.548 versus AIR's 0.679, a 13.1-point gap attributable to the adversarial probing architecture rather than model capacity. Second, gpt-oss-20b is a mixture-of-experts model with $\sim$3.6B active parameters per forward pass, comparable to or smaller than IntentRL's backbone in active compute, yet AIR with gpt-oss-20b still exceeds IntentRL's schema-only F1 (0.626 vs.\ 0.618). Together, these comparisons indicate that AIR's advantage comes from the adversarial framework itself rather than backbone scale.

\subsubsection{Performance with Full Database as Context}

Table~\ref{tab:overall} also shows performance when the full database is available as context.  IntentRL leads with higher metrics (recall 0.854, F1 0.786) than AIR with Claude Sonnet 4.5 (recall 0.818, F1 0.729). However, this comes at the cost of generating 5 predictions per query regardless of ambiguity, of which only 2.99 are unique execution outputs—meaning $\sim$40\% of IntentRL's generations are duplicates. AIR's adversarial probing and execution-based pruning ensure all outputs are unique, generating only 1.76 per query on average. More predicted candidates mean more chances to match a gold output, but this also comes at a direct cost to user experience.  In a production Text-to-SQL system, each additional interpretation requires a follow-up clarification from the user. IntentRL produces extra outputs on 336 of 440 queries (76.4\%), while AIR does so on only 145 (33.0\%).  IntentRL would require clarification on more than double the queries. Table~\ref{tab:generation} summarizes the generation efficiency.

\begin{table}[t]
\centering
\small
\setlength{\tabcolsep}{4pt}
\begin{tabular}{lcc}
\toprule
\textbf{Metric} & \textbf{AIR} & \textbf{IntentRL} \\
\midrule
Avg.\ output / query & 1.76 & 5.00 \\
Avg.\ unique outputs / query & 1.76 & 2.99 \\
Extra outputs ($>$ gold) & 145 (33.0\%) & 336 (76.4\%) \\
\bottomrule
\end{tabular}
\caption{Comparison of SQL execution outputs between AIR and IntentRL on AMBROSIA with the full database as context. }
\label{tab:generation}
\end{table}

\subsubsection{Robustness: Schema-Only vs.\ Full Database}
When moving from full database to schema-only context, AIR (Claude Sonnet 4.5) degrades by only $\sim$7\% across recall, precision, and F1, while IntentRL degrades by $\sim$19--22\% on the same metrics. This $\sim$3$\times$ gap reveals that IntentRL has learned to rely on actual data in the database to generate correct SQL. AIR's adversarial architecture, which reasons at the intent level before committing to SQL, maintains its effectiveness regardless of whether actual data in the database is available or not. Combined with its training-free design, this makes AIR practical for privacy-sensitive and latency-constrained production environments where schema-only access is the only option.

\subsubsection{Perfect Match Accuracy}

The recall, precision, and F1 metrics above use partial credit: e.g., a prediction covering 3 of 4 gold cell values counts as a partial match. This can obscure whether a system produces usable outputs in practice. In a deployed data conversation system, a user should only be asked for clarification when there is genuine ambiguity, and the system's interpretation set should perfect match the true set of valid interpretations, no missing readings, no extra ones. We therefore calculate perfect match accuracy, a metric with two requirements: (1)~every gold execution output is exactly reproduced by some prediction (using the same cell-multiset comparison as All-Found), and (2)~the number of unique predicted outputs equals the number of gold outputs, penalizing over-generation. A query is correct only if both conditions hold.

\begin{table}[t]
\centering
\small
\setlength{\tabcolsep}{5pt}
\begin{tabular}{lcc}
\toprule
\textbf{Context} & \textbf{AIR (Claude Sonnet 4.5)} & \textbf{IntentRL} \\
\midrule
Schema only   & 204 (46.4\%) & 102 (23.2\%) \\
Full database & 216 (49.1\%) & 89 (20.2\%) \\
\bottomrule
\end{tabular}
\caption{Perfect match accuracy on AMBROSIA. A query is correct only if all gold outputs are perfectly matched and no extra outputs are produced. Under schema-only context, AIR retains 94.4\% of the best full database performance without requiring actual data for intent discovery or SQL generation.}
\label{tab:exact}
\end{table}

Table~\ref{tab:exact} shows results under both context settings. Under schema-only context, AIR achieves 204 perfect matches, maintaining 94.4\% of the best full database performance (216) without requiring access to actual data in the database to discover intents and generate SQL queries. This demonstrates that AIR's adversarial probing can identify the correct interpretation set from schema information alone. In contrast, IntentRL's fixed 5-candidate strategy produces more unique outputs than gold interpretations on 76\% of queries, failing condition~(2) even when all gold outputs are covered (273 of 362 All-Found queries fail due to extra outputs). AIR's targeted generation tends to trigger clarification only when genuinely necessary. Examples of AIR failure cases are provided in Appendix~\ref{app:errors}.

\subsubsection{Threshold Sensitivity Analysis}

\begin{figure}[t]
\centering
\includegraphics[width=\columnwidth]{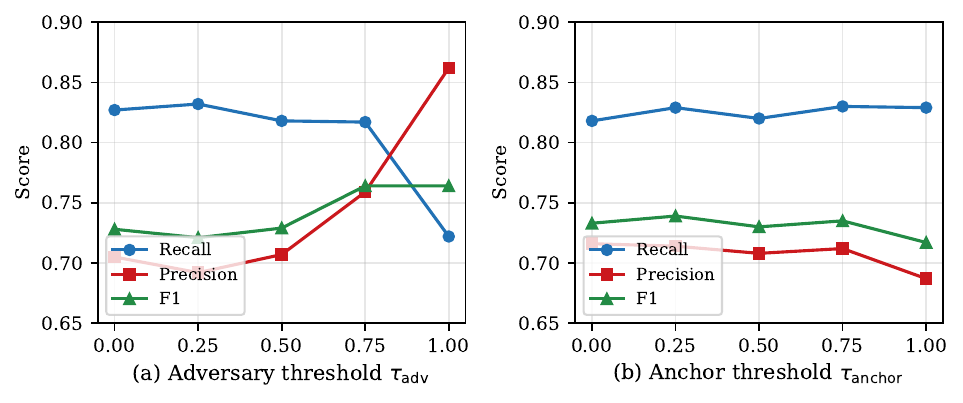}
\caption{Effect of confidence thresholds on AMBROSIA with the full database as context. (a)~The adversary threshold $\tau_{\text{adv}}$ controls a clear recall--precision tradeoff: higher values filter more aggressively, improving precision at the cost of recall. (b)~The anchor threshold $\tau_{\text{anchor}}$ has minimal effect, with all metrics varying within $\sim$3 percentage points.}
\label{fig:threshold}
\end{figure}

Figure~\ref{fig:threshold} shows how AIR's two confidence thresholds affect the recall--precision tradeoff on AMBROSIA with the full database as context.

The adversary confidence threshold $\tau_{\text{adv}}$ (Figure~\ref{fig:threshold}(a)) is the primary tuning parameter. At $\tau_{\text{adv}} = 0$ (accept all adversary alternatives), recall peaks at 0.827 but precision is lowest (0.705). As the threshold increases, low-confidence alternatives are filtered: at $\tau_{\text{adv}} = 0.75$, precision rises to 0.759 while recall remains competitive (0.817), yielding the highest F1 (0.764). It should be noted that, at $\tau_{\text{adv}} = 1.0$, all adversary alternatives are rejected, reducing AIR to a single-interpretation system: precision reaches 0.862 but recall drops to 0.722. The 10.5\% recall gap between $\tau_{\text{adv}} = 0$ and $\tau_{\text{adv}} = 1$ confirms that adversarial probing contributes meaningful coverage, while the adversary's self-reported confidence provides a useful signal for filtering implausible alternatives.

This tradeoff has an asymmetric cost in deployment: missing a genuine ambiguity silently returns a possibly wrong answer, while a false alarm only costs one extra clarification round. This asymmetry favors operating points that trade some precision for recall, i.e., lower values of $\tau_{\text{adv}}$ than the default. All main results in this paper use $\tau_{\text{adv}} = 0.5$ as a balanced default; deployments where missed ambiguity is the more costly failure mode should lower $\tau_{\text{adv}}$.

The anchor confidence threshold $\tau_{\text{anchor}}$ (Figure~\ref{fig:threshold}b) has minimal impact: all metrics vary within $\sim$3 percentage points across the full range. This indicates that the adversary threshold alone suffices to control output quality. All main results in this paper use $\tau_{\text{adv}} = 0.5$ and $\tau_{\text{anchor}} = 0.95$.

\subsection{Cross-Benchmark Transferability}

Table~\ref{tab:ambiqt} presents results on AmbiQT, where ambiguity arises from schema design (synonym columns/tables, split tables, pre-aggregated views) rather than linguistic structure. This benchmark tests whether AIR's adversarial probing generalizes beyond the linguistic ambiguity types.

\begin{table*}[t]
\centering
\small
\setlength{\tabcolsep}{8pt}
\begin{tabular}{llccccc}
\toprule
\textbf{Context} & \textbf{Method} & \textbf{Recall} & \textbf{Precision} & \textbf{F1} & \textbf{One-Found} & \textbf{All-Found} \\
\midrule
\multirow{2}{*}{Full database}
 & IntentRL   & \textbf{0.695} & 0.590 & 0.594 & \textbf{0.866} & \textbf{0.525} \\
 & AIR -- Claude Sonnet 4.5 (ours) & 0.650 & \textbf{0.758} & \textbf{0.686} & 0.862 & 0.439 \\
\addlinespace[4pt]
\multirow{2}{*}{Schema only}
 & IntentRL   & \textbf{0.684} & 0.549 & 0.570 & \textbf{0.842} & \textbf{0.527} \\
 & AIR -- Claude Sonnet 4.5 (ours) & 0.642 & \textbf{0.743} & \textbf{0.675} & 0.847 & 0.436 \\
\bottomrule
\end{tabular}
\caption{Performance on AmbiQT validation set to test cross-benchmark generalization. AIR under schema-only achieves near-identical performance to its full database setting. AIR achieves higher F1 through substantially better precision, while IntentRL's exhaustive generation yields higher recall and All-Found.}
\label{tab:ambiqt}
\end{table*}

AIR achieves higher F1 under both context settings (0.686 vs.\ 0.594 full database; 0.675 vs.\ 0.570 schema-only), driven by substantially higher precision (0.758 vs.\ 0.590). IntentRL achieves higher recall (0.695 vs.\ 0.650) and All-Found (0.525 vs.\ 0.439) through generating more predictions per query. Notably, AIR under schema-only (F1 0.675) achieves near-identical performance to its full database  setting (F1 0.686). This confirms that schema-level ambiguities are inherently context-independent and that AIR's adversarial probing does not rely on actual data. 

\section{Discussion}

\paragraph{Why schema-only matters.}
In production deployments, exposing full database contents to an LLM raises privacy, latency, and cost concerns. This concern is specifically about what is sent to the LLM as context; ``schema-only'' governs intent discovery and SQL generation, not SQL execution, which still runs agains full database under standard database access control. AIR's adversarial probing operates at the intent level, reasoning about what a query \emph{could mean} rather than what the data contains, which explains why removing data has low impact ($\sim$7\% degradation). IntentRL's larger degradation ($\sim$20\%) appears to stem from data-dependent shortcuts acquired during RL fine-tuning. On AmbiQT, where IntentRL was not trained, both methods degrade $<$4\%.

\paragraph{Why adversarial probing outperforms both prompting and RL.}
Prompting and RL represent two ends of a spectrum, but both treat interpretation discovery as a single-pass problem. Prompting for ``multiple interpretations'' produces surface-level variants because the model's probability mass concentrates around its most likely reading (prompt baseline recall: 0.697 vs.\ AIR's 0.818). RL rewards diversity but without verifying that candidates are genuinely distinct, leading to over-generation. AIR solves both: the negative constraint forces the Adversary to explore \emph{away} from the Anchor's reading, and execution-based pruning retains only candidates with provably different outputs.

\paragraph{Targeted generation vs.\ fixed-budget coverage.}
IntentRL generates 5 candidates per query regardless of ambiguity, producing extra outputs on 76\% of queries with each risking an unnecessary user clarification. AIR generates alternatives only when a plausible competing interpretation is found (1.76 outputs on average). The adversary threshold $\tau_{\text{adv}}$ provides a tuning knob: lowering it increases coverage at the cost of more clarifications; raising it reduces interruptions at the cost of missing some ambiguities.

\paragraph{Execution as proof of ambiguity.}
Ambiguity in AIR is not merely detected but \emph{proven}: two interpretations are genuinely ambiguous only when their SQL translations produce different results. Under schema-only context, execution still provides a divergence signal through structural differences (different joins, column selections, or filtering logic), explaining why AIR's precision remains high (0.658 vs.\ 0.589 for IntentRL) despite the absence of data content.

\paragraph{Evaluation granularity.}
The execution-based metrics rely on execution output matching between predicted and gold outputs. This penalizes predictions that are reasonable from a user's perspective but differ in surface form. For instance, AIR may include an identifying column alongside the requested data (e.g., returning \texttt{labPosition, salaryRange} when only \texttt{salaryRange} is gold), or interpret ``accommodate 500 people'' as $\geq$500 rather than exactly 500. Both are plausible choices that receive no credit under strict output matching. These cases indicate that execution output matching does not distinguish between SQL generation errors and legitimate interpretation differences, potentially understating practical correctness. An LLM-as-judge evaluation that assesses equivalence at multiple levels including intent coverage, SQL structural correctness, and output semantics could provide a more faithful measure of system quality for ambiguity-aware Text-to-SQL.

\section{Conclusion}

We presented AIR, a multi-agent framework for ambiguity-aware Text-to-SQL that requires no task-specific training. AIR detects ambiguity through adversarial probing, verifies it through execution-based divergence, and resolves it through contrastive clarification.

Under schema-only context, the realistic production setting where privacy, cost, and latency constraints prevent data exposure, AIR outperforms the RL-trained IntentRL across all metrics (e.g., F1 0.679 vs.\ 0.618), with roughly one-third the performance degradation when full database is unavailable ($\sim$7\% vs.\ $\sim$20\%). On perfect match accuracy, AIR under schema-only retains 94.4\% of the best full-database performance without requiring actual data for intent discovery or SQL generation. On AmbiQT, AIR achieves higher F1 without any benchmark-specific adaptation, demonstrating cross-benchmark transferability.

AIR's targeted generation (1.76 outputs per query) tends to trigger clarification only when a genuine competing interpretation is found, in contrast to IntentRL's fixed 5-candidate strategy that produces extra outputs on 76\% of queries. These results show that structured adversarial reasoning at inference time offers a practical alternative to training-based approaches for ambiguity-aware Text-to-SQL, with superior robustness, precision, and deployability.

\clearpage
\section*{Limitations}

AIR requires multiple LLM calls per query (Anchor, Adversary, SQL generation for each interpretation, and Synthesizer), resulting in $\sim$\ 40 seconds end-to-end latency per query using unoptimized serial calls to the Bedrock Claude Sonnet 4.5 API. The approximate per-stage median breakdown is: Anchor $\sim$9s, Adversary $\sim$22s, SQL generation $\sim$6s, Synthesizer $\sim$5s. The Adversary dominates due to its multi-round reasoning constrained by prior interpretations; even the Anchor alone takes $\sim$9s, likely reflecting API network latency rather than model computation. Optimizations such as parallelization of agent calls and caching would reduce this for real-time production deployment.

Both benchmarks we evaluate on are single-turn; measuring cumulative user burden and task completion across multi-turn clarification dialogues, and extending evaluation to larger real-world Text-to-SQL benchmarks such as BIRD once multi-interpretation annotations are available, are directions for future work.

While we evaluate whether the correct interpretation set is produced, we do not evaluate the quality of the generated clarification questions from a user perspective.  A human evaluation or LLM-as-judge assessment of their naturalness, informativeness, and usefulness to end users is left to future work.

\section*{Ethical Considerations}
This work does not create any new datasets. All experiments use existing public benchmarks  (AMBROSIA and AmbiQT) that have been verified to contain no sensitive or harmful information.

\bibliography{custom}

\appendix

\section{Agent Prompts}
\label{app:prompts}

This appendix presents the core prompts (implemented as DSPy \citep{khattab2024dspy}) for AIR's three main agents. All agents share a common set of few-shot examples covering scope, attachment, and vagueness ambiguity types (see Section~\ref{subsec:few-shot}).

\subsection{Anchor Agent Prompt}

\begin{quote}
\small
\textbf{System:} Identify the single most likely interpretation of the user's query against the database schema. Analyze the question step-by-step, map terms to schema elements, and commit to one clear interpretation. State your mapping choices --- do NOT discuss alternative interpretations or what you didn't choose.

\medskip
\textbf{Inputs:}
\begin{itemize}
    \item \texttt{query}: The user's natural language query
    \item \texttt{database\_dump}: Database schema and sample data
\end{itemize}

\textbf{Outputs:}
\begin{itemize}
    \item \texttt{interpretation}: Step-by-step analysis: parse the question, map each key term to table.column in the schema, identify any ambiguity, state your interpretation.
    \item \texttt{confidence}: Confidence 0.0--1.0 that this is the only reasonable interpretation.
\end{itemize}
\end{quote}

\subsection{Adversary Agent Prompt}

\begin{quote}
\small
\textbf{System:} Determine if the query has a genuinely different interpretation from the anchor's. A valid alternative interpretation must be completely supported by the natural language query without hallucination.

\medskip
\textbf{Inputs:}
\begin{itemize}
    \item \texttt{query}: The user's natural language query
    \item \texttt{database\_dump}: Database schema and sample data
    \item \texttt{anchor\_interpretation}: The anchor's interpretation of the query
\end{itemize}

\textbf{Outputs:}
\begin{itemize}
    \item \texttt{found\_alternative}: true if a valid but genuinely different interpretation exists, false otherwise.
    \item \texttt{alternative\_interpretation}: Step-by-step analysis for the alternative interpretation. Write `NONE' if no alternative exists.
    \item \texttt{alternative\_confidence}: Confidence 0.0--1.0 that the alternative is valid. 0.0 if no alternative.
\end{itemize}
\end{quote}

\subsection{SQL Generation Agent Prompt}

\begin{quote}
\small
\textbf{System:} Generate a SQL query from the interpretation. Translate the interpretation into valid SQLite SQL. Output ONLY raw SQL, no markdown fences. SELECT only the columns the question asks for.

\medskip
\textbf{Inputs:}
\begin{itemize}
    \item \texttt{query}: The user's natural language query
    \item \texttt{database\_dump}: Database schema and sample data
    \item \texttt{interpretation}: Step-by-step interpretation mapping question terms to schema elements
\end{itemize}

\textbf{Outputs:}
\begin{itemize}
    \item \texttt{sql\_query}: Raw SQL query implementing the interpretation.
\end{itemize}
\end{quote}

\section{AIR Error Examples}
\label{app:errors}

The following examples illustrate representative failure modes where AIR generates incorrect outputs.

\paragraph{Extra interpretations.}
For the query ``Provide a list of all horror films and also list thriller films with a budget of 20 million. Show them in one table,'' the ground truth applies the budget filter only to thrillers:

\smallskip
\noindent\texttt{\small SELECT title WHERE genre = 'Horror' OR (genre = 'Thriller' AND budget = 20M)} $\to$ 3 rows
\smallskip

\noindent AIR's Adversary generates an additional interpretation where the budget applies to both genres:

\smallskip
\noindent\texttt{\small SELECT title WHERE (genre = 'Horror' OR genre = 'Thriller') AND budget = 20M} $\to$ 2 rows
\smallskip

\paragraph{Extra columns.}
For ``What is the salary range for female laboratory technicians and research scientists? Show them in one table,'' the ground truth selects only \texttt{salaryRange}, but AIR selects \texttt{labPosition, salaryRange}, adding an identifying column to disambiguate the salary ranges.

\paragraph{Extra rows.}
For ``List all lobbies or terraces that can accomodate 500 people. Show them in one table,'' the ground truth applies the size constraint to both types:

\smallskip
\noindent\texttt{\small SELECT Name WHERE type IN ('Lobby','Terrace') AND Size = 500} $\to$ 2 rows
\smallskip

\noindent AIR interprets ``accommodate 500'' as $\geq$500 rather than exactly 500:

\smallskip
\noindent\texttt{\small SELECT Name WHERE type = 'Lobby' AND Size >= 500 OR type = 'Terrace'} $\to$ 4 rows
\smallskip

\paragraph{Missing interpretations.}
For ``What experience do I need for the open position with the greatest salary?'' the schema has both \texttt{Minimum\_Years} and \texttt{Preferred\_Years} columns. The ground truth expects three interpretations (each column separately, plus both). AIR generates only:

\smallskip
\noindent\texttt{\small SELECT Minimum\_Years FROM Jobs ORDER BY Salary DESC LIMIT 1}
\smallskip

\noindent The Adversary does not recognize that ``experience'' is vague across the two columns and fails to propose the alternative mappings.

\section*{License}
The AMBROSIA dataset is used under its original license as provided by \citet{saparina2025ambrosia}. The AmbiQT dataset is used under its original license as provided by \citet{bhaskar-etal-2023-benchmarking}. All open source models (Gemma 4 26B A4B, gpt-oss-20b, IntentRL) are used under their original licenses.

\section*{Computational Budget}                                                                
AIR is training-free and requires only inference-time computation. The total experimental budget spans 3 backbone models, 2 context settings on AMBROSIA (440 samples), a prompt baseline, and 9 threshold sweep configurations. Additionally, 2 context settings on AmbiQT (1,911 samples).

Claude Sonnet 4.5 was accessed via Amazon Bedrock API. Open source models, gpt-oss-20b and IntentRL, were run internally on A100 GPU nodes. Total cumulative inference time was approximately 200 GPU-hours across all experiments.

\end{document}